\documentclass[conference]{IEEEtran}

\usepackage{cite}
\usepackage{amsmath,amssymb,amsfonts}
\usepackage{graphicx}
\usepackage{textcomp}
\usepackage{xcolor}
\def\BibTeX{{\rm B\kern-.05em{\sc i\kern-.025em b}\kern-.08em
    T\kern-.1667em\lower.7ex\hbox{E}\kern-.125emX}}

\usepackage{multirow}
\usepackage{tabularx}
\usepackage{xspace}
\usepackage{enumitem}

\usepackage{algorithm}
\PassOptionsToPackage{noend}{algpseudocode}
\usepackage{algpseudocode}
\errorcontextlines\maxdimen

\usepackage{tikz}
\usepackage{amsmath}
\usepackage{makecell}

\newcommand{\sol}{{\em FusedInf}\xspace}

\makeatletter
\def\ps@IEEEtitlepagestyle{%
  \def\@oddfoot{\mycopyrightnotice}%
  \def\@evenfoot{}%
}

\def\mycopyrightnotice{%
  {\footnotesize \textcolor{red}{\begin{tabular}[t]{@{}l@{}} This paper has been accepted for publication by the 9th ACM/IEEE Symposium on Edge Computing (SEC). © 2024 IEEE. Personal use of this material is\\permitted. Permission from IEEE must be obtained for all other uses, in any current or future media, including reprinting/republishing this material for\\advertising or promotional purposes, creating new collective works, for resale or redistribution to servers or lists, or reuse of any copyrighted component of\\this work in other works.\end{tabular}}}
  \gdef\mycopyrightnotice{}
}

\begin{document}

\title{\sol: Efficient Swapping of DNN Models for On-Demand Serverless Inference Services on the Edge}

\author{
\IEEEauthorblockN{Sifat Ut Taki}
\IEEEauthorblockA{University of Notre Dame\\
staki@nd.edu}
\and
\IEEEauthorblockN{Arthi Padmanabhan}
\IEEEauthorblockA{Harvey Mudd College\\
arpadmanabhan@g.hmc.edu}
\and
\IEEEauthorblockN{Spyridon Mastorakis}
\IEEEauthorblockA{University of Notre Dame \\
mastorakis@nd.edu}
}

\maketitle

\begin{abstract}
Edge AI computing boxes are a new class of computing devices that are aimed to revolutionize the AI industry. These compact and robust hardware units bring the power of AI processing directly to the source of data--on the edge of the network. On the other hand, on-demand serverless inference services are becoming more and more popular as they minimize the infrastructural cost associated with hosting and running DNN models for small to medium-sized businesses. However, these computing devices are still constrained in terms of resource availability. As such, the service providers need to load and unload models efficiently in order to meet the growing demand. In this paper, we introduce \sol to efficiently swap DNN models for on-demand serverless inference services on the edge. \sol combines multiple models into a single Direct Acyclic Graph (DAG) to efficiently load the models into the GPU memory and make execution faster. Our evaluation of popular DNN models showed that creating a single DAG can make the execution of the models up to 14\% faster while reducing the memory requirement by up to 17\%. The prototype implementation is available at https://github.com/SifatTaj/FusedInf.
\end{abstract}


\section{Introduction}

\begin{table*}
    \caption{Examples of commercial AI edge computing boxes.}
    \label{tab:edge-box}
    \centering
    \small
    \begin{tabular}{|c|c|c|c|c|} \hline
         \textbf{Vendor} &\textbf{Model}&  \textbf{CPU}&  \textbf{GPU}&  \textbf{Memory}\\ \hline 
         VVDN&Xavier NX &  \makecell{6-core NVIDIA\\Carmel ARM v8.2}&  \makecell{NVIDIA Volta architecture\\with 384 NVIDIA CUDA cores\\and 48 Tensor cores}&  8GB\\ \hline 
          Advantech&EPC-R7300IJ&  8-core ARM Cortex&  \makecell{NVIDIA Ampere GPU\\with 1024 CUDA cores\\and 32 Tensor Cores}&  16GB\\ \hline 
          FORLINX&FCU3001&  \makecell{6-core NVIDIA\\Carmel ARM v8.2}&  \makecell{NVIDIA Volta architecture\\with 384 NVIDIA CUDA cores\\and 48 Tensor cores}&  16GB\\ \hline
 EdgeMatrix& Advance Mk2& \makecell{9th/8th Gen\\Intel Core i7}& \makecell{NVIDIA Tesla T4\\with 2560 CUDA core\\and 320 Tensor cores}&16GB\\\hline
 Azure Edge Stack& Pro 2& \makecell{Intel Xeon\\Gold 6209U}& \makecell{NVIDIA A2 with\\1280 CUDA cores and\\40 Tensor cores}&32GB\\\hline
    \end{tabular}
    \vspace{-0.3cm}
\end{table*}

Efficient DNN inference is crucial for making deep learning models practical in real-world applications on the edge~\cite{mcenroe2022survey, kamruzzaman2021new}. While DNNs are extremely capable of tasks like image recognition and speech translation, running them often requires significant computing power. This can be a bottleneck for deploying them on resource-constrained devices or in situations demanding fast response times. By optimizing DNN inference to use less power and run efficiently, we can unlock the potential of deep learning for a wider range of applications--from medical diagnosis on mobile devices to real-time obstacle detection in autonomous vehicles on edge.

\begin{figure}
    \centering
    \includegraphics[width=\linewidth]{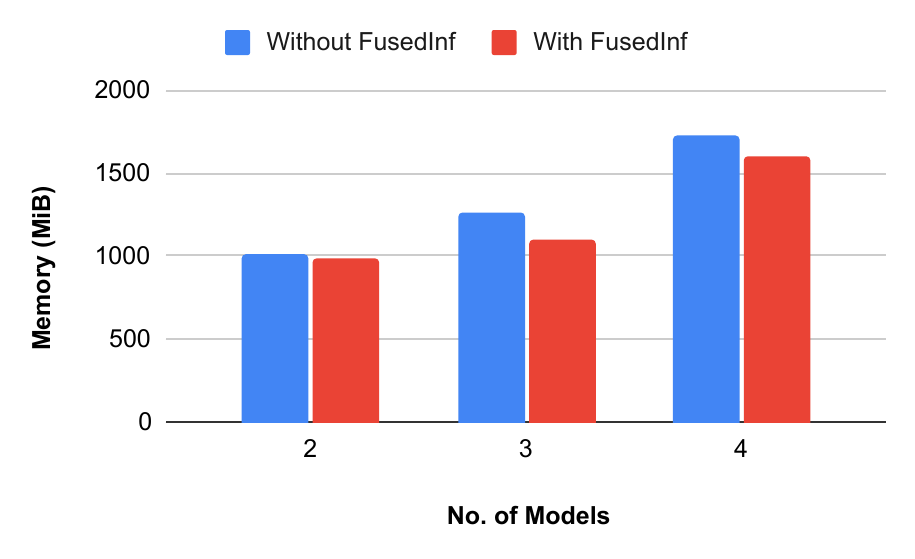}
    \vspace{-0.4cm}
    \caption{Effectiveness of \sol when concurrently executing multiple DNN models for inference on the edge.}
    \vspace{-0.3cm}
    \label{fig:intro_result}
\end{figure}

Various cloud service providers are allowing users to deploy and run their models on the cloud and edge. However, owning and managing a virtual machine on the cloud for inference is very expensive--especially for small businesses, which do not always require the inference service. Hence, it is not economically viable for them to keep an inference service running idle. As a solution, cloud providers have developed serverless inference services for on-demand inference~\cite{jarachanthan2021amps, yang2022infless}. This makes it easier to deploy and run DNN models; however, it severely complicates the process on the service provider's end--especially on the edge where resources are limited. The service providers need to constantly swap models on the limited resources of the edge boxes. Swapping models comes at the cost of an overhead of loading and unloading them every time a new model is queried by a user.

in order to address the aforementioned problem, we need to come up with a system that efficiently swaps models on edge devices. However, we should address a few challenges to ensure the security and correctness of each user and model. (C1) Can we ensure model correctness while retaining the accuracy of the models? (C2) Will the system efficiently work for a wide variety of models? (C3) How much overhead will there be? (C4) Can it be achieved while ensuring the privacy of the users?

\sol introduces a novel approach to efficiently swap models on the edge with very limited overhead while addressing the challenges mentioned above. \sol compiles a unified directed acyclic graph (DAG) of multiple models before loading them on the GPU memory. This facilitates the process of loading and querying multiple models at the same time. Figure~\ref{fig:intro_result} presents an overview of peak GPU memory usage when executing VGG16~\cite{vgg}, MobileNetv3~\cite{mobilenets}, DenseNet161~\cite{densenet}, and EfficientNetv2Large~\cite{efficientnet} for inference with and without \sol. The following are the major contributions of this paper:

\begin{itemize}
    \item We discuss the challenges and opportunities of optimizing CUDA operations for DNN model executions on edge.
    
    \item We introduce a multi-model DAG compilation technique to compile graphs for efficient loading of multiple models into the GPU memory.

    \item We demonstrate the effectiveness of this technique through a comprehensive evaluation of multiple DNN models.
\end{itemize}

\section{Background}

The convergence of machine learning (ML) and edge computing is driving the development of new, powerful technologies specifically designed for edge boxes~\cite{yao2022edge, firouzi2022convergence, wu2020cloud}. These compact computing devices process data locally--at the source of generation much closer to the users--rather than relying on communication with the cloud. This trend necessitates a paradigm shift from traditional cloud-optimized DNN models to lightweight and resource-efficient models tailored for resource-constrained edge devices~\cite{zhou2019edge}. One promising area of research is in the development of optimized AI models, which achieve acceptable accuracy with minimal computational power~\cite{guo2021mistify}. Additionally, the field is witnessing the emergence of frameworks specifically designed for edge deployment, focusing on optimizing AI workloads for edge hardware and enabling real-time, on-device inferencing. This synergy between AI and edge computing holds immense potential for applications across various sectors, from industrial automation and predictive maintenance to intelligent traffic management and personalized healthcare.

DNNs are revolutionizing various fields due to their ability to solve complex problems. However, training and running these computationally intensive models requires significant processing power. Compute Unified Device Architecture (CUDA)~\cite{cuda} plays a vital role in accelerating DNN models by leveraging the parallel processing capabilities of Graphics Processing Units (GPUs).  CUDA provides a programming model and a set of development tools that allow developers to write DNN algorithms in languages like C++, exploiting the massive core count of GPUs for tasks like matrix multiplication and convolution, which are fundamental operations in DNNs.  Furthermore, libraries like CUDA Deep Neural Network library (cuDNN~\cite{cudnn}) offer highly optimized implementations of commonly used DNN primitives, which further accelerate training and inference processes. The use of CUDA in DNN model training has significantly reduced training times and enabled the development of ever-more complex and powerful neural networks.

Within the NVIDIA ecosystem, cuDNN and CUDA APIs play significant roles in developing and deploying complex models. Their significance lies in two primary areas: performance optimization and developer efficiency. cuDNN leverages the parallel processing capabilities of NVIDIA GPUs, providing highly optimized implementations for fundamental deep learning operations like convolution, pooling, and activation functions. This offloading of computationally intensive tasks from CPUs to GPUs translates to significant speedups in training and inference times compared to CPU-only implementations. This performance boost is essential for training large-scale models with billions of parameters within reasonable timeframes. Furthermore, cuDNN simplifies the development process by offering pre-optimized kernels that integrate seamlessly with popular deep-learning frameworks like TensorFlow and PyTorch. This abstraction layer allows researchers and engineers to focus on model design and experimentation without going into low-level GPU programming, accelerating innovation in the field. As such, cuDNN's performance optimizations and developer-friendly interface make it an indispensable tool for researchers and engineers pushing the boundaries of deep learning.

\subsection{CUDA API Operations}

One critical aspect to consider is the distinction between host (CPU) and device (GPU) memory. DNN models consist of weights, biases, and activation layers, all requiring memory storage.  CUDA provides mechanisms for allocating memory on the GPU using functions like \textit{cudaMalloc()} and \textit{cudaMemcpy()}. However, simply allocating sufficient memory isn't enough as Fragmentation (where allocated memory becomes scattered across the GPU memory space) can occur, which may hinder performance. Frameworks often employ memory pools to mitigate this issue, allocating contiguous memory blocks for better utilization~\cite{gao2020estimating}.

Another key factor influencing memory usage is the data flow during DNN execution. Forward and backward passes in training involve numerous intermediate tensors representing activations and gradients. While some frameworks like cuDNN offer optimizations to reduce memory footprint, these intermediate tensors still consume significant resources. Techniques like checkpointing, where intermediate states are periodically saved to host memory, can be employed to free up GPU memory.

Furthermore, the choice of data type for model parameters significantly impacts memory requirements. While single-precision floating-point numbers (FP32) offer high accuracy, they can be memory-intensive for large models. Techniques like mixed-precision training, where computations are performed using lower precision formats like FP16, can significantly reduce memory usage without sacrificing substantial accuracy~\cite{mixedprec}.

Beyond model parameters and intermediate tensors, other factors contribute to GPU memory consumption. Frameworks themselves allocate memory for internal data structures and workspace for cuDNN operations (such as CUDA CONTEXT). Moreover, depending on the complexity of the DNN architecture, additional memory might be required for storing activation functions and gradients.

Optimizing memory allocation for DNNs on GPUs requires a holistic approach. Techniques like memory profiling tools can help identify memory bottlenecks and guide optimization efforts. Utilizing techniques like gradient accumulation, where gradients for multiple mini-batches are accumulated before updating weights, can reduce memory requirements for backpropagation. Additionally, exploring alternative DNN architectures with lower memory footprints might be necessary for resource-constrained GPUs.

\begin{table}
\caption{Comparison of different memory/latency saving frameworks.}
\centering 
\label{tab:frame_comp}
\small
\begin{tabular}{|l|l|l|l|}
\hline
\textbf{Framework} & \textbf{Application} & \textbf{\makecell[l]{Requires\\model\\similarity}} & \textbf{\makecell[l]{Impacts\\accuracy}} \\ \hline
Gemel~\cite{gemel}              & \makecell[l]{Video\\analytics}      & Partial                                    & Yes                      \\ \hline
HiveMind~\cite{hivemind}           & \makecell[l]{Concurrent\\training}  & Full                                       & Yes                      \\ \hline
Mainstream~\cite{mainstream}         & \makecell[l]{Video\\analytics}      & Partial                                    & Yes                      \\ \hline
AdaShare~\cite{adashare}           & \makecell[l]{Multi-task\\learning}  & Partial                                    & Yes                      \\ \hline
FusedInf           & \makecell[l]{Serverless\\inference} & No                                         & No                       \\ \hline
\end{tabular}
\vspace{-0.4cm}
\end{table}

\begin{figure*}
    \centering
    \includegraphics[width=\textwidth]{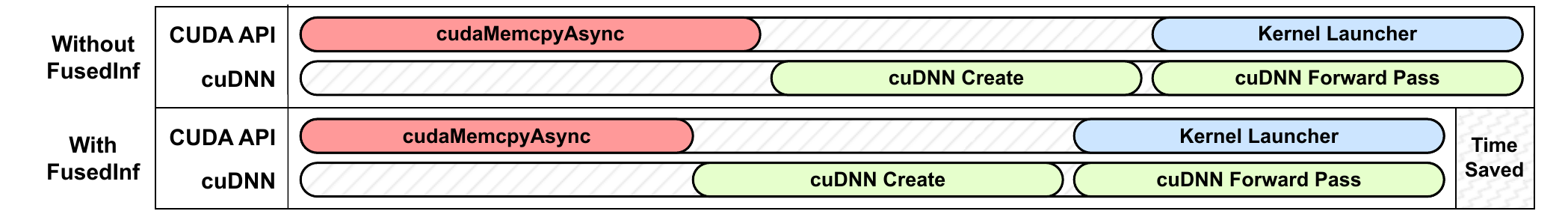}
    \vspace{-0.4cm}
    \caption{DNN model execution timeline on a GPU.}
    \vspace{-0.4cm}
    \label{fig:timeline}
\end{figure*}

\subsection{Recent Advancements}
Model merging and operator fusion can optimize the concurrent execution of multiple DNN models on the edge. Gemel introduces a technique called model merging to improve memory usage on edge devices for real-time video analytics on the edge. It merges similar layers from different models to reduce the overall memory footprint and the time it takes to swap data between host and GPU memory.

HiveMind, on the other hand, takes a similar approach to Gemel. HiveMind is a system designed to speed up the concurrent execution of multiple DNN models. It achieves this by grouping models into batches, then performing operator fusion across these models and sharing data efficiently. It utilizes a parallel processing system to execute this optimized group of models for faster performance. However, HiveMind requires manual model grouping. Similar to Gemel, HiveMind performs cross-model layer fusion when stateful operators in different models share the same underlying weights or when stateful operators have the same input and output shapes. This is an unlikely scenario in a serverless inference service where different users will query different models at a time. 

Table~\ref{tab:frame_comp} presents a comparison between our proposed approach and other DNN memory/latency saving frameworks. The aforementioned approaches save memory or time by sharing or reusing model layers and operators across multiple DNN models. This requires architectural similarity among those models. However, an on-demand serverless inference service on the edge may need to run a wide variety of models from different users that may contain little to no architectural similarity. Additionally, model similarity search introduces a significant overhead, which is not tolerable for fast and efficient model swapping on an edge device. Moreover, sharing layers across multiple models comes at a cost of reduced accuracy. As a service provider, it is important to ensure model correctness for serverless inference services as users expect no accuracy degradation during inference.

\section{Challenges \& Motivation}

With the emergence of on-demand serverless inference services on the edge, service providers are expected to face a massive amount of traffic querying different models throughout the day~\cite{cisco}. However, commercial edge boxes are extremely resource-contained compared to cloud servers. Table~\ref{tab:edge-box} presents some of the commercially available edge AI computing boxes today. These edge boxes rely on NVIDIA CUDA technology for DNN computations.

\subsection{CUDA Optimization Challenges}

Optimizing the CUDA operations for serverless inference services is challenging as it is important to make sure the users are served correctly on time. In order to make a robust and efficient serverless inference system on edge, the following challenges need to be addressed:

\noindent\textbf{C1: Retaining model correctness:} As a service provider, it is important to ensure that the user-provided DNN models perform \textit{exactly} the way it is meant to be. As a result, DNN model architecture-level optimizations are limited as techniques like layer merging and mixed precision execution are not feasible because they will impact the correctness of the individual DNN models.

\noindent\textbf{C2: Compatibility with model variations:} It is expected that a service provider will be receiving requests from a wide variety of users--each querying different DNN models with different weights. As such, optimizations like operator fusion among different models will not be possible as it requires the DNN models to have the same layer architecture, weights, and input shape.

\noindent\textbf{C3: Minimal overhead:} A serverless inference system performs queries on a DNN model for a limited amount of time before the models need to be swapped to serve another set of users. This swapping operation might need to be performed 100-1000 times a day depending on the users' demand and traffic load for a specific edge box. As such, any optimization that incurs a significant overhead cannot be employed in a serverless inference service on edge.

\noindent\textbf{C4: Ensuring privacy:} When performing inference on a DNN model using user input data, ensuring privacy is essential. Optimizations like operator fusion and layer merging can leak data from one model to another--violating user privacy and leaving a massive security vulnerability in the system.

\subsection{Motivation}

\begin{figure}
    \centering
    \includegraphics[width=\linewidth]{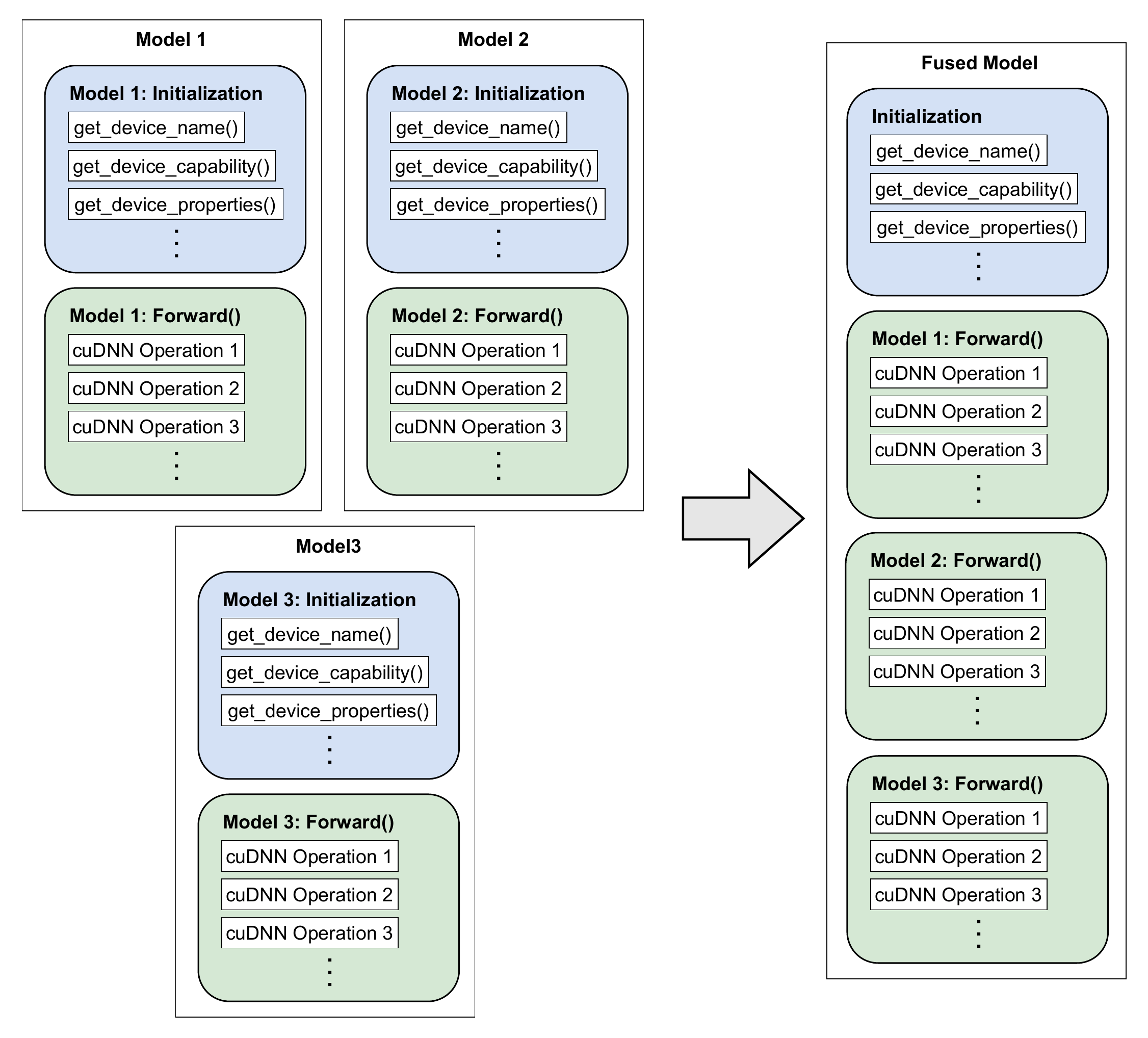}
    \caption{Efficiency in function invocation when compiling a single DAG of multiple models.}
    \vspace{-0.4cm}
    \label{fig:fn_call}
\end{figure}

In order to build an effective system that can swap models efficiently on edge devices, we need to address the challenges discussed in the previous section. When executing multiple DNN models, the number of core DNN operations (operations on each layer) should remain the same to ensure that each model produces the expected output. For example, the number of convolution operations and linear operations should remain the same when executing different computer vision DNN models. However, there are other function calls being made when loading the models into the GPU memory. Functions that are responsible for loading the libraries, initiating the models, configuring the devices, etc. As such, optimizations can be done when loading the models by eliminating a few of the redundant function calls. Subsequently, these optimizations can be generalized to all types DNN models since every DNN model initialization follows a similar set of function calls. As a result, there is no need for the models to be similar in order to optimize the model execution. Figure~\ref{fig:fn_call} presents an overview of the function calls when initializing multiple DNN models. If a single DAG is compiled with multiple model architectures, the functions needed to initialize the model will be called only once. This should make the model initialization process more efficient, which is crucial as the DNN models need to be initialized every time when swapping models. Moreover, segmented memory allocation could be inefficient when initializing models separately. This process can also be facilitated if a single DAG is compiled and initialized. So, faster memory allocations should also be possible by compiling the graphs efficiently. Finally, these should result in higher throughput--allowing more data to be moved in a given time. All of these optimizations should make model swapping more efficient on an edge device.

\section{System Design}

In this section, we first discuss the overall design of \sol and how it operates on an edge node for serverless inference. Next, we discuss how \sol optimizes the CUDA operations while addressing the optimization challenges for efficient swapping of models on edge devices.

\subsection{System Architecture}

\begin{figure}
    \centering
    \includegraphics[width=0.8\linewidth]{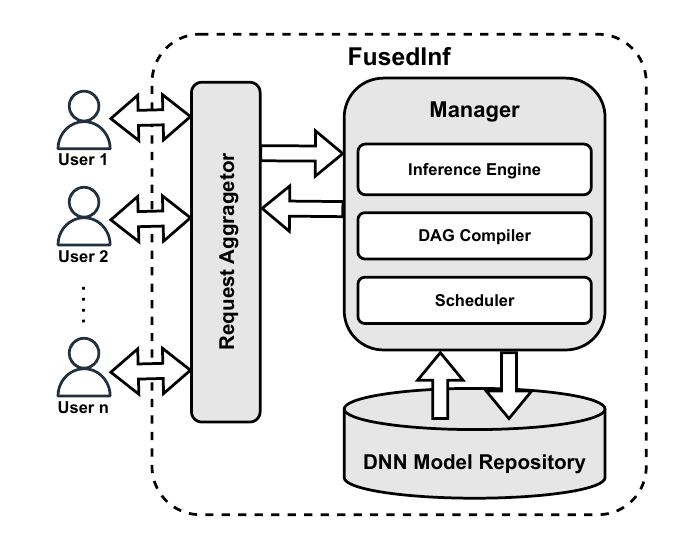}
    \vspace{-0.3cm}
    \caption{\sol system architecture.}
    \label{fig:sys_arch}
\end{figure}

We developed a prototype of \sol, which can be deployed on commercial edge AI boxes. This framework is expected to be deployed on edge boxes handling thousands of requests a day querying a wide variety of DNN models. The framework has the following components:

\noindent\textbf{DNN model repository}: The DNN model repository stores the model architectures and the trained weights for all the DNN models on that particular edge AI computing box. User-registered DNN models are offloaded to the closest edge AI box and stored in the DNN model repository for performing inference in the future. Since the framework is for serverless inference services on edge, DNN model architectures and the associated parameters need to be loaded on the GPU memory whenever there's a request from the user of that particular model. Along with the DNN model architectures and weights, the model repository also stores the GPU memory requirement and the inference latency for each DNN model.

\noindent\textbf{Request aggregator}: The request aggregator validates and aggregates all the upcoming valid queries from different users and forwards them to the manager. Considering it is deployed in a highly demanding scenario, the request aggregator keeps aggregating the upcoming requests while the system is occupied processing current requests.

\begin{figure}
    \centering
    \includegraphics[width=\linewidth]{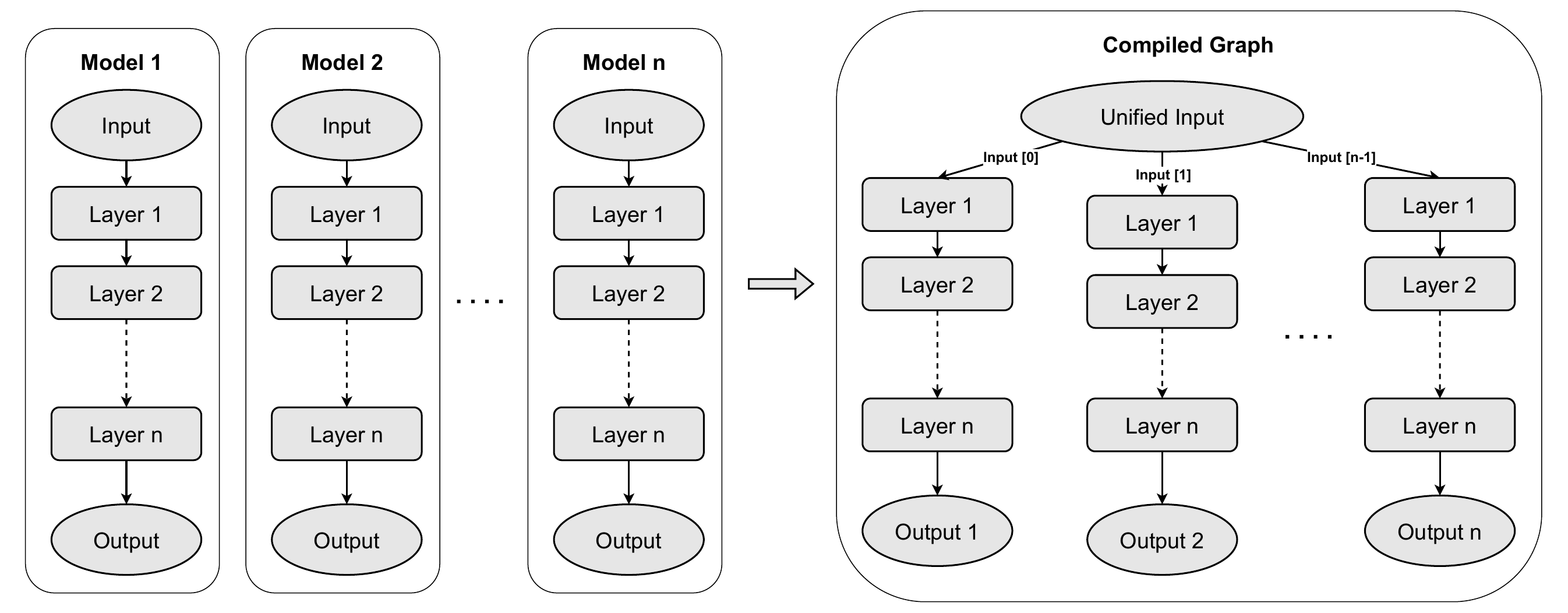}
    \vspace{-0.5cm}
    \caption{Graph compilation process of \sol with multiple DNN models.}
    \vspace{-0.4cm}
    \label{fig:dag}
\end{figure}

\noindent\textbf{Manager}: Manager is the core of the framework. It is responsible for controlling the entire system. It has three sub-components: a DAG compiler, an inference engine, and a scheduler. Depending on the aggregated requests, the manager determines how many models the DAG compiler can compile into a DAG. Since the DNN model repository stores the memory requirement information for each DNN model, the manager can estimate the number of DNN models to compile depending on the available GPU memory on the system. Once it is determined, the DAG compiler retrieves the DNN model architectures and weights from the DNN model repository and compiles the graph as demonstrated in Figure~\ref{fig:dag}. Subsequently, the compiled DAG is forwarded to the inference engine where it utilizes the GPU for running the inferences using Algorithm~\ref{algo:dag_comp}.

\sol is vertically scalable with multiple GPUs. The Manager is capable of compiling and scheduling multiple DAGs at a time, keeping the resources occupied. When scheduling, \sol takes the model uptime into account and groups the models with short-term requests together. For example, models that require a single inference will be grouped together while models that require longer runtime will be grouped separately when compiling multiple DAGs. \sol is also capable of dynamically alerting a compiled DAG to swap a sub-graph if needed. When a sub-graph within the DAG requires swapping, it can be recompiled efficiently with the remaining sub-graphs.

The framework facilitates the swapping of DNN models on an edge device. Subsequently, \sol is suitable for scheduling on resource-constrained edge devices. For example, when an edge box is required to run more DNN models than it can fit in its GPU memory for a long period of time, \sol can batch the DNN models and efficiently swap the DAGs periodically. For long-term executions of multiple DNN models, the models are grouped into batches such that each batch can fit into the GPU memory. Subsequently, each batch of models is compiled into a corresponding DAG. The scheduler runs a fixed number of iterations (set by the service provider) and swaps it with the next DAG--following a round-robin scheduling algorithm.

\begin{algorithm}
\caption{\sol Manager}\label{algo}
\label{algo:dag_comp}
\renewcommand{\algorithmicrequire}{\textbf{Input:}}
\renewcommand{\algorithmicensure}{\textbf{Output:}}
\begin{algorithmic}[1]
\Require DNN models $\{m_1, m_2, \dotsc, m_n\}\in \mathcal{M}$, inputs $\{x_1, x_2, \dotsc, x_n\}\in \mathcal{X}$

\Ensure Predictions $\mathcal{Y}$

\Statex \textit{Initialization} : $\mathcal{G} \leftarrow \emptyset$, $\mathcal{Y} \leftarrow \emptyset$

\For{$\forall m \in \mathcal{M}$}
    \For{$u_i, u_j \in m$}
        \State $\mathcal{G} \leftarrow \{u_i^m\}_{i=1}^n, \{u_i^m, u_j^m\}, \{\phi_i^m\}_{i=1}^n$
    \EndFor
\EndFor

\For{$\forall x^m \in \mathcal{X}$}
    \For{$\forall u \in \mathcal{G}$}
        \State $\mathcal{Y} \leftarrow \phi_i^m(\sum_{j=1}^{L_i} u_{i,j}^{m} x_{i,j}^m)$
    \EndFor
\EndFor
\Return $\mathcal{Y}$
\end{algorithmic}
\end{algorithm}

\subsection{\sol Optimization Techniques}
\sol addresses the aforementioned challenges to achieve efficiency when swapping DNN models on an edge AI computing device using the following techniques.

\noindent\textbf{Single process execution:} When executing multiple DNN models at a time, creating separate processes is inefficient. When separate processes are invoked for each DNN model, they all require individual loading of essential libraries to execute the model in a GPU. As a result, it creates a significant overhead when loading a DNN model into GPU memory in terms of execution time and memory consumption. CUDA Multi-Process Service (MPS) is supposed to help with the process; however, it is still inefficient and unreliable~\cite{hivemind}. \sol tackles this problem by creating a single process for all the models--eliminating the redundancy of loading the necessary libraries separately for each process.

\noindent\textbf{Faster model initialization:} When initializing multiple DNN models, some CUDA functions are redundantly called for each model. For example, functions like \textit{cuDeviceGet()} and \textit{cudaGetDevice()} are used to get device information and architecture compatibility. Other functions like \textit{get\_schema()} are called to fetch the schema of the model being initiated, \textit{cudaGetDeviceCount()} is called to get the number of CUDA devices, \textit{cuDriverGet()} to fetch driver information, etc. \sol eliminates the redundant calls by compiling and initializing a single DAG, which results in faster model initialization. Table~\ref{tab:fn_calls} presents different function calls and their execution time with 7 different DNN models.

\noindent\textbf{Fewer memory calls:} When a DNN model is executed on a GPU, data needs to be moved from the host to the GPU. \textit{cudaMemcpyAsync()} is a function in the CUDA Runtime API that allows transfers of data between host and device memory asynchronously. This enables the program to continue execution while the data transfer happens in the background, potentially improving overall performance. The function can be optionally linked to a specific CUDA stream, which helps manage the order of data transfers on the GPU. \sol optimizes this function call by compiling a single DAG of multiple models--resulting in fewer calls of this function. As a result, it facilitates the loading of models on edge devices.

\noindent\textbf{Efficient memory allocation:} \textit{cudaMalloc()} is a function used in CUDA programming to allocate memory on the GPU. This function allows requests for a specific amount of space in GPU memory and then provides a pointer to that memory location. This pointer can then be used to transfer data to the GPU memory and perform computations on that data. By compiling a single DAG, \sol makes this memory allocation significantly faster. Our experiment with 7 DNN models suggests that \sol can make this operation 90\% quicker and makes 12\% fewer calls that save memory.

\noindent\textbf{Higher throughput:} By compiling a single DAG, \sol achieves higher throughput. Our experiment with 7 DNN models showed 1.44 GiB/s higher throughput. Figure~\ref{fig:timeline} presents the CUDA operation timeline of DNN model execution on a GPU.

\begin{table}
\caption{Breakdown of different CUDA functions and their execution time when initialized 7 DNN models.}
\label{tab:fn_calls}
\small
\centering
\begin{tabular}{|l|l|l|c|}
\hline
\textbf{CUDA Function}     & \textbf{\makecell{Time\\without\\ \sol}} & \textbf{\makecell{Time\\with\\ \sol}} & \textbf{Decrease} \\ \hline
cuDeviceGet           & 740 ns                                & 500 ns                               & \textbf{32.4\%}      \\ \hline
cuDeviceGetCount      & 921 ns                                & 701 ns                               & \textbf{23.9\%}      \\ \hline
cuDriverGetVersion    & 251 ns                                & 100 ns                               & \textbf{60.2\%}      \\ \hline
cudaGetDevice         & 4.12 ms                               & 4.01 ms                              & \textbf{2.7\%}       \\ \hline
cudaGetDeviceCount    & 440 ns                                & 390 ns                               & \textbf{11.4\%}      \\ \hline
cudaMalloc            & 443.8 ms                              & 40.1 ms                              & \textbf{91.0\%}      \\ \hline
cudaMemcpyAsync       & 2.964 s                               & 2.759 s                              & \textbf{6.9\%}       \\ \hline
cudaSetDevice         & 2.54 ms                               & 2.50 ms                              & \textbf{1.6\%}       \\ \hline
cudaStreamIsCapturing & 75.40 ms                              & 72.36 ms                             & \textbf{4.0\%}       \\ \hline
\end{tabular}
\vspace{-0.3cm}
\end{table}

\subsection{How \sol Addresses the Challenges}
Adopting the aforementioned optimizations, \sol can facilitate the DNN model swapping operation--allowing a service provider to serve more users per day in highly demanding scenarios. 

\begin{itemize}
    \item \sol addresses the challenge of retaining model correctness (C1) by not altering the model architectures when compiling the DAG. The compiled DAG consists of sub-graphs of each model exactly the way a user had provided containing the exact number of DNN operators. As a result, the output of each model remains the same.

    \item The DAG compiler can compile any DNN model architecture, which ensures compatibility with model variations (C2). Section~\ref{sec:eval} presents the evaluation of \sol with a wide variety of model architectures to show the compatibility and adaptability of \sol in different applications. Moreover, \sol works with all types of DNNs since it does not depend on the individual model architecture or architectural similarities across fused models. It is designed to optimize CUDA functionality when initializing multiple models. Fundamentally, every DNN model initialization follows a similar set of function calls. As a result, the speed-up does not depend on any specific combination of models, and we have not noticed any slowdown with any model combinations.

    \item \sol leverages fast and efficient DAG compilation for minimal overhead. Moreover, it makes fewer memory calls, faster memory allocation, and achieves higher throughput. As a result, it addresses the challenge of achieving a minimal overhead (C3).

    \item Finally, \sol does not fuse or merge operators from models across. Each sub-graph processes its own input without sharing outputs from the layers, which addresses the challenge of ensuring privacy (C4).
\end{itemize}

\section{Evaluation}
\label{sec:eval}

\begin{table*}
\caption{DNN models used for different test cases.}
\label{teb:test_cases}
\centering
\begin{tabular}{|ll|cc|llllllllll|}
\hline
\multicolumn{2}{|c|}{\multirow{2}{*}{\textbf{Model}}}                      & \multicolumn{2}{c|}{\textbf{Accuracy}}               & \multicolumn{10}{c|}{\textbf{Test Cases}}                                                                                                                                                                                                                                                                                                          \\ \cline{3-14} 
\multicolumn{2}{|c|}{}                                                     & \multicolumn{1}{c|}{\textbf{Top 1}} & \textbf{Top 5} & \multicolumn{1}{c|}{\textbf{1}} & \multicolumn{1}{c|}{\textbf{2}} & \multicolumn{1}{c|}{\textbf{3}} & \multicolumn{1}{c|}{\textbf{4}} & \multicolumn{1}{c|}{\textbf{5}} & \multicolumn{1}{c|}{\textbf{6}} & \multicolumn{1}{c|}{\textbf{7}} & \multicolumn{1}{c|}{\textbf{8}} & \multicolumn{1}{c|}{\textbf{9}} & \multicolumn{1}{c|}{\textbf{10}} \\ \hline
\multicolumn{1}{|l|}{AlexNet}                      & AlexNet               & \multicolumn{1}{c|}{56.522}         & 79.066         & \multicolumn{1}{l|}{}           & \multicolumn{1}{c|}{\textbf{X}} & \multicolumn{1}{c|}{\textbf{X}} & \multicolumn{1}{l|}{\textbf{}}  & \multicolumn{1}{l|}{}           & \multicolumn{1}{l|}{}           & \multicolumn{1}{c|}{\textbf{X}} & \multicolumn{1}{l|}{}           & \multicolumn{1}{c|}{\textbf{X}} &                                  \\ \hline
\multicolumn{1}{|l|}{\multirow{8}{*}{VGG}}         & VGG-11                & \multicolumn{1}{c|}{69.02}          & 88.628         & \multicolumn{1}{l|}{}           & \multicolumn{1}{l|}{}           & \multicolumn{1}{l|}{}           & \multicolumn{1}{c|}{\textbf{X}} & \multicolumn{1}{l|}{}           & \multicolumn{1}{l|}{}           & \multicolumn{1}{l|}{}           & \multicolumn{1}{l|}{}           & \multicolumn{1}{c|}{\textbf{X}} & \textbf{}                        \\ \cline{2-14} 
\multicolumn{1}{|l|}{}                             & VGG-13                & \multicolumn{1}{c|}{69.928}         & 89.246         & \multicolumn{1}{l|}{}           & \multicolumn{1}{l|}{}           & \multicolumn{1}{l|}{}           & \multicolumn{1}{l|}{}           & \multicolumn{1}{l|}{}           & \multicolumn{1}{l|}{}           & \multicolumn{1}{l|}{}           & \multicolumn{1}{l|}{\textbf{}}  & \multicolumn{1}{l|}{}           & \multicolumn{1}{c|}{\textbf{X}}  \\ \cline{2-14} 
\multicolumn{1}{|l|}{}                             & VGG-16                & \multicolumn{1}{c|}{71.592}         & 90.382         & \multicolumn{1}{l|}{}           & \multicolumn{1}{l|}{\textbf{}}  & \multicolumn{1}{l|}{}           & \multicolumn{1}{l|}{}           & \multicolumn{1}{l|}{}           & \multicolumn{1}{l|}{\textbf{}}  & \multicolumn{1}{l|}{}           & \multicolumn{1}{c|}{\textbf{X}} & \multicolumn{1}{l|}{}           & \textbf{}                        \\ \cline{2-14} 
\multicolumn{1}{|l|}{}                             & VGG-19                & \multicolumn{1}{c|}{72.376}         & 90.876         & \multicolumn{1}{l|}{}           & \multicolumn{1}{c|}{\textbf{X}} & \multicolumn{1}{l|}{}           & \multicolumn{1}{l|}{}           & \multicolumn{1}{l|}{}           & \multicolumn{1}{c|}{\textbf{X}} & \multicolumn{1}{l|}{\textbf{}}  & \multicolumn{1}{l|}{}           & \multicolumn{1}{l|}{}           & \multicolumn{1}{c|}{\textbf{X}}  \\ \cline{2-14} 
\multicolumn{1}{|l|}{}                             & VGG-11 with batch normalization        & \multicolumn{1}{c|}{70.37}          & 89.81          & \multicolumn{1}{l|}{}           & \multicolumn{1}{l|}{}           & \multicolumn{1}{l|}{\textbf{}}  & \multicolumn{1}{l|}{}           & \multicolumn{1}{l|}{}           & \multicolumn{1}{l|}{}           & \multicolumn{1}{c|}{\textbf{X}} & \multicolumn{1}{l|}{}           & \multicolumn{1}{l|}{}           &                                  \\ \cline{2-14} 
\multicolumn{1}{|l|}{}                             & VGG-13 with batch normalization        & \multicolumn{1}{c|}{71.586}         & 90.374         & \multicolumn{1}{l|}{\textbf{}}  & \multicolumn{1}{l|}{}           & \multicolumn{1}{c|}{\textbf{X}} & \multicolumn{1}{l|}{}           & \multicolumn{1}{l|}{}           & \multicolumn{1}{l|}{}           & \multicolumn{1}{l|}{}           & \multicolumn{1}{l|}{}           & \multicolumn{1}{l|}{}           &                                  \\ \cline{2-14} 
\multicolumn{1}{|l|}{}                             & VGG-16 with batch normalization        & \multicolumn{1}{c|}{73.36}          & 91.516         & \multicolumn{1}{c|}{\textbf{X}} & \multicolumn{1}{l|}{}           & \multicolumn{1}{l|}{}           & \multicolumn{1}{l|}{\textbf{}}  & \multicolumn{1}{l|}{}           & \multicolumn{1}{l|}{}           & \multicolumn{1}{l|}{}           & \multicolumn{1}{l|}{}           & \multicolumn{1}{l|}{}           &                                  \\ \cline{2-14} 
\multicolumn{1}{|l|}{}                             & VGG-19 with batch normalization        & \multicolumn{1}{c|}{74.218}         & 91.842         & \multicolumn{1}{l|}{\textbf{}}  & \multicolumn{1}{l|}{}           & \multicolumn{1}{l|}{}           & \multicolumn{1}{c|}{\textbf{X}} & \multicolumn{1}{l|}{}           & \multicolumn{1}{l|}{\textbf{}}  & \multicolumn{1}{l|}{}           & \multicolumn{1}{l|}{}           & \multicolumn{1}{l|}{}           &                                  \\ \hline
\multicolumn{1}{|l|}{\multirow{5}{*}{ResNet}}      & ResNet-18             & \multicolumn{1}{c|}{69.758}         & 89.078         & \multicolumn{1}{c|}{\textbf{X}} & \multicolumn{1}{l|}{}           & \multicolumn{1}{l|}{\textbf{}}  & \multicolumn{1}{l|}{\textbf{}}  & \multicolumn{1}{l|}{\textbf{}}  & \multicolumn{1}{c|}{\textbf{X}} & \multicolumn{1}{l|}{}           & \multicolumn{1}{l|}{}           & \multicolumn{1}{l|}{\textbf{}}  &                                  \\ \cline{2-14} 
\multicolumn{1}{|l|}{}                             & ResNet-34             & \multicolumn{1}{c|}{73.314}         & 91.42          & \multicolumn{1}{l|}{}           & \multicolumn{1}{l|}{\textbf{}}  & \multicolumn{1}{c|}{\textbf{X}} & \multicolumn{1}{c|}{\textbf{X}} & \multicolumn{1}{c|}{\textbf{X}} & \multicolumn{1}{l|}{\textbf{}}  & \multicolumn{1}{l|}{}           & \multicolumn{1}{l|}{}           & \multicolumn{1}{c|}{\textbf{X}} &                                  \\ \cline{2-14} 
\multicolumn{1}{|l|}{}                             & ResNet-50             & \multicolumn{1}{c|}{76.13}          & 92.862         & \multicolumn{1}{l|}{}           & \multicolumn{1}{c|}{\textbf{X}} & \multicolumn{1}{l|}{}           & \multicolumn{1}{l|}{}           & \multicolumn{1}{l|}{}           & \multicolumn{1}{c|}{\textbf{X}} & \multicolumn{1}{l|}{\textbf{}}  & \multicolumn{1}{l|}{\textbf{}}  & \multicolumn{1}{l|}{}           &                                  \\ \cline{2-14} 
\multicolumn{1}{|l|}{}                             & ResNet-101            & \multicolumn{1}{c|}{77.374}         & 93.546         & \multicolumn{1}{l|}{}           & \multicolumn{1}{l|}{}           & \multicolumn{1}{l|}{}           & \multicolumn{1}{l|}{\textbf{}}  & \multicolumn{1}{l|}{}           & \multicolumn{1}{l|}{}           & \multicolumn{1}{c|}{\textbf{X}} & \multicolumn{1}{c|}{\textbf{X}} & \multicolumn{1}{l|}{}           &                                  \\ \cline{2-14} 
\multicolumn{1}{|l|}{}                             & ResNet-152            & \multicolumn{1}{c|}{78.312}         & 94.046         & \multicolumn{1}{l|}{}           & \multicolumn{1}{l|}{}           & \multicolumn{1}{l|}{}           & \multicolumn{1}{c|}{\textbf{X}} & \multicolumn{1}{l|}{}           & \multicolumn{1}{l|}{}           & \multicolumn{1}{l|}{\textbf{}}  & \multicolumn{1}{c|}{\textbf{X}} & \multicolumn{1}{l|}{}           &                                  \\ \hline
\multicolumn{1}{|l|}{\multirow{2}{*}{SqueezeNet}}  & SqueezeNet 1.0        & \multicolumn{1}{c|}{58.092}         & 80.42          & \multicolumn{1}{l|}{}           & \multicolumn{1}{l|}{}           & \multicolumn{1}{l|}{}           & \multicolumn{1}{l|}{}           & \multicolumn{1}{l|}{\textbf{}}  & \multicolumn{1}{l|}{}           & \multicolumn{1}{c|}{\textbf{X}} & \multicolumn{1}{l|}{}           & \multicolumn{1}{l|}{}           &                                  \\ \cline{2-14} 
\multicolumn{1}{|l|}{}                             & SqueezeNet 1.1        & \multicolumn{1}{c|}{58.178}         & 80.624         & \multicolumn{1}{l|}{}           & \multicolumn{1}{l|}{}           & \multicolumn{1}{l|}{}           & \multicolumn{1}{l|}{}           & \multicolumn{1}{c|}{\textbf{X}} & \multicolumn{1}{l|}{}           & \multicolumn{1}{l|}{}           & \multicolumn{1}{l|}{}           & \multicolumn{1}{l|}{}           &                                  \\ \hline
\multicolumn{1}{|l|}{\multirow{4}{*}{DenseNet}}    & Densenet-121          & \multicolumn{1}{c|}{74.434}         & 91.972         & \multicolumn{1}{l|}{}           & \multicolumn{1}{l|}{}           & \multicolumn{1}{l|}{}           & \multicolumn{1}{l|}{}           & \multicolumn{1}{l|}{}           & \multicolumn{1}{l|}{}           & \multicolumn{1}{l|}{}           & \multicolumn{1}{l|}{}           & \multicolumn{1}{l|}{}           &                                  \\ \cline{2-14} 
\multicolumn{1}{|l|}{}                             & Densenet-169          & \multicolumn{1}{c|}{75.6}           & 92.806         & \multicolumn{1}{l|}{}           & \multicolumn{1}{l|}{}           & \multicolumn{1}{l|}{\textbf{}}  & \multicolumn{1}{l|}{}           & \multicolumn{1}{l|}{}           & \multicolumn{1}{l|}{}           & \multicolumn{1}{l|}{}           & \multicolumn{1}{l|}{}           & \multicolumn{1}{l|}{}           &                                  \\ \cline{2-14} 
\multicolumn{1}{|l|}{}                             & Densenet-201          & \multicolumn{1}{c|}{76.896}         & 93.37          & \multicolumn{1}{l|}{\textbf{}}  & \multicolumn{1}{l|}{}           & \multicolumn{1}{c|}{\textbf{X}} & \multicolumn{1}{l|}{}           & \multicolumn{1}{l|}{}           & \multicolumn{1}{l|}{}           & \multicolumn{1}{l|}{}           & \multicolumn{1}{l|}{}           & \multicolumn{1}{l|}{}           &                                  \\ \cline{2-14} 
\multicolumn{1}{|l|}{}                             & Densenet-161          & \multicolumn{1}{c|}{77.138}         & 93.56          & \multicolumn{1}{c|}{\textbf{X}} & \multicolumn{1}{l|}{}           & \multicolumn{1}{l|}{}           & \multicolumn{1}{l|}{}           & \multicolumn{1}{l|}{}           & \multicolumn{1}{l|}{}           & \multicolumn{1}{l|}{}           & \multicolumn{1}{l|}{}           & \multicolumn{1}{l|}{}           &                                  \\ \hline
\multicolumn{1}{|l|}{Inception}                    & Inception v3          & \multicolumn{1}{c|}{77.294}         & 93.45          & \multicolumn{1}{l|}{}           & \multicolumn{1}{l|}{}           & \multicolumn{1}{l|}{}           & \multicolumn{1}{l|}{}           & \multicolumn{1}{l|}{}           & \multicolumn{1}{l|}{}           & \multicolumn{1}{l|}{}           & \multicolumn{1}{l|}{}           & \multicolumn{1}{l|}{}           &                                  \\ \hline
\multicolumn{1}{|l|}{GoogLeNet}                    & GoogLeNet             & \multicolumn{1}{c|}{69.778}         & 89.53          & \multicolumn{1}{l|}{}           & \multicolumn{1}{l|}{}           & \multicolumn{1}{l|}{\textbf{}}  & \multicolumn{1}{l|}{\textbf{}}  & \multicolumn{1}{l|}{}           & \multicolumn{1}{l|}{}           & \multicolumn{1}{l|}{}           & \multicolumn{1}{l|}{}           & \multicolumn{1}{l|}{}           &                                  \\ \hline
\multicolumn{1}{|l|}{\multirow{2}{*}{ShuffleNet}}  & ShuffleNet V2 x1.0    & \multicolumn{1}{c|}{69.362}         & 88.316         & \multicolumn{1}{l|}{}           & \multicolumn{1}{l|}{}           & \multicolumn{1}{c|}{\textbf{X}} & \multicolumn{1}{c|}{\textbf{X}} & \multicolumn{1}{l|}{\textbf{}}  & \multicolumn{1}{l|}{}           & \multicolumn{1}{l|}{}           & \multicolumn{1}{l|}{}           & \multicolumn{1}{l|}{}           &                                  \\ \cline{2-14} 
\multicolumn{1}{|l|}{}                             & ShuffleNet V2 x0.5    & \multicolumn{1}{c|}{60.552}         & 81.746         & \multicolumn{1}{l|}{\textbf{}}  & \multicolumn{1}{l|}{}           & \multicolumn{1}{l|}{}           & \multicolumn{1}{l|}{}           & \multicolumn{1}{c|}{\textbf{X}} & \multicolumn{1}{l|}{\textbf{}}  & \multicolumn{1}{l|}{}           & \multicolumn{1}{l|}{}           & \multicolumn{1}{l|}{}           & \textbf{}                        \\ \hline
\multicolumn{1}{|l|}{\multirow{3}{*}{MobileNet}}   & MobileNetV2           & \multicolumn{1}{c|}{71.878}         & 90.286         & \multicolumn{1}{c|}{\textbf{X}} & \multicolumn{1}{l|}{\textbf{}}  & \multicolumn{1}{l|}{}           & \multicolumn{1}{l|}{}           & \multicolumn{1}{c|}{\textbf{X}} & \multicolumn{1}{c|}{\textbf{X}} & \multicolumn{1}{l|}{}           & \multicolumn{1}{l|}{}           & \multicolumn{1}{l|}{}           & \multicolumn{1}{c|}{\textbf{X}}  \\ \cline{2-14} 
\multicolumn{1}{|l|}{}                             & MobileNet V3 Large    & \multicolumn{1}{c|}{74.042}         & 91.34          & \multicolumn{1}{l|}{}           & \multicolumn{1}{c|}{\textbf{X}} & \multicolumn{1}{l|}{}           & \multicolumn{1}{l|}{}           & \multicolumn{1}{l|}{}           & \multicolumn{1}{l|}{}           & \multicolumn{1}{l|}{}           & \multicolumn{1}{l|}{}           & \multicolumn{1}{l|}{}           & \multicolumn{1}{c|}{\textbf{X}}  \\ \cline{2-14} 
\multicolumn{1}{|l|}{}                             & MobileNet V3 Small    & \multicolumn{1}{c|}{67.668}         & 87.402         & \multicolumn{1}{l|}{}           & \multicolumn{1}{l|}{}           & \multicolumn{1}{l|}{}           & \multicolumn{1}{l|}{}           & \multicolumn{1}{l|}{}           & \multicolumn{1}{l|}{}           & \multicolumn{1}{l|}{}           & \multicolumn{1}{l|}{}           & \multicolumn{1}{l|}{}           & \textbf{}                        \\ \hline
\multicolumn{1}{|l|}{\multirow{2}{*}{ResNeXt}}     & ResNeXt-50-32x4d      & \multicolumn{1}{c|}{77.618}         & 93.698         & \multicolumn{1}{l|}{}           & \multicolumn{1}{l|}{}           & \multicolumn{1}{l|}{}           & \multicolumn{1}{l|}{}           & \multicolumn{1}{l|}{}           & \multicolumn{1}{l|}{}           & \multicolumn{1}{l|}{\textbf{}}  & \multicolumn{1}{l|}{\textbf{}}  & \multicolumn{1}{l|}{}           & \multicolumn{1}{c|}{\textbf{X}}  \\ \cline{2-14} 
\multicolumn{1}{|l|}{}                             & ResNeXt-101-32x8d     & \multicolumn{1}{c|}{79.312}         & 94.526         & \multicolumn{1}{l|}{}           & \multicolumn{1}{l|}{\textbf{}}  & \multicolumn{1}{l|}{}           & \multicolumn{1}{l|}{}           & \multicolumn{1}{l|}{}           & \multicolumn{1}{l|}{\textbf{}}  & \multicolumn{1}{c|}{\textbf{X}} & \multicolumn{1}{c|}{\textbf{X}} & \multicolumn{1}{l|}{\textbf{}}  &                                  \\ \hline
\multicolumn{1}{|l|}{\multirow{2}{*}{Wide ResNet}} & Wide ResNet-50-2      & \multicolumn{1}{c|}{78.468}         & 94.086         & \multicolumn{1}{l|}{}           & \multicolumn{1}{c|}{\textbf{X}} & \multicolumn{1}{l|}{}           & \multicolumn{1}{l|}{}           & \multicolumn{1}{l|}{\textbf{}}  & \multicolumn{1}{c|}{\textbf{X}} & \multicolumn{1}{l|}{}           & \multicolumn{1}{l|}{}           & \multicolumn{1}{c|}{\textbf{X}} &                                  \\ \cline{2-14} 
\multicolumn{1}{|l|}{}                             & Wide ResNet-101-2     & \multicolumn{1}{c|}{78.848}         & 94.284         & \multicolumn{1}{l|}{}           & \multicolumn{1}{l|}{}           & \multicolumn{1}{l|}{}           & \multicolumn{1}{l|}{}           & \multicolumn{1}{c|}{\textbf{X}} & \multicolumn{1}{l|}{}           & \multicolumn{1}{l|}{}           & \multicolumn{1}{l|}{\textbf{}}  & \multicolumn{1}{l|}{\textbf{}}  &                                  \\ \hline
\multicolumn{1}{|l|}{\multirow{2}{*}{MNASNet}}     & MNASNet 1.0           & \multicolumn{1}{c|}{73.456}         & 91.51          & \multicolumn{1}{l|}{}           & \multicolumn{1}{l|}{}           & \multicolumn{1}{l|}{}           & \multicolumn{1}{l|}{}           & \multicolumn{1}{l|}{}           & \multicolumn{1}{l|}{}           & \multicolumn{1}{l|}{}           & \multicolumn{1}{c|}{\textbf{X}} & \multicolumn{1}{c|}{\textbf{X}} &                                  \\ \cline{2-14} 
\multicolumn{1}{|l|}{}                             & MNASNet 0.5           & \multicolumn{1}{c|}{67.734}         & 87.49          & \multicolumn{1}{l|}{\textbf{}}  & \multicolumn{1}{l|}{}           & \multicolumn{1}{l|}{}           & \multicolumn{1}{l|}{}           & \multicolumn{1}{l|}{}           & \multicolumn{1}{l|}{}           & \multicolumn{1}{l|}{}           & \multicolumn{1}{l|}{}           & \multicolumn{1}{l|}{}           &                                  \\ \hline
\multicolumn{1}{|l|}{EfficientNet}                 & EfficientNet V2 Large & \multicolumn{1}{c|}{85.808}         & 97.788         & \multicolumn{1}{c|}{\textbf{X}} & \multicolumn{1}{l|}{}           & \multicolumn{1}{l|}{}           & \multicolumn{1}{l|}{}           & \multicolumn{1}{l|}{}           & \multicolumn{1}{l|}{}           & \multicolumn{1}{l|}{}           & \multicolumn{1}{l|}{}           & \multicolumn{1}{l|}{}           &                                  \\ \hline
\end{tabular}
\end{table*}

In order to evaluate our system, we selected a few popular DNN models. We picked 34 different models from 13 different model families. We ran our evaluations in three phases. In phase one, we started with one model and gradually increased the number of models up to 7 to see the impact of combining models. In the next phase, we swapped 5 models randomly for 10 different consecutive test cases. For both phases, we ran 100 iterations for each model and collected the peak GPU memory consumption and the total execution time. The execution time metric was an average of 5 runs, which included the model loading time and the time to run 100 iterations. In the third phase, we evaluated the dynamic sub-graph swapping capability of \sol by swapping a sub-graph after a certain iteration for 5 different sub-graphs in a DAG. The system used for evaluation consisted of an AMD Ryzen Threadripper 5955WX and an NVIDIA RTX 4090. Table~\ref{teb:test_cases} presents the models used for different test cases for the second phase along with their ImageNet-1K accuracy. Following are the brief descriptions of the models used for evaluation.

We used VGG models~\cite{vgg}, short for Visual Geometry Group models, which are a family of convolutional neural networks (CNNs) known for their simplicity and effectiveness in image recognition tasks. We used AlexNet~\cite{alexnet}--named after Alex Krizhevsky, which is a convolutional neural network (CNN) architecture that revolutionized image recognition in 2012. We also picked ResNet~\cite{resnet} (short for Residual Neural Network)--a deep learning architecture specifically designed to address the vanishing gradient problem that can hinder training in very deep neural networks. Furthermore, we picked SqueezeNet~\cite{squeezenet}, which is designed for efficiency. Unlike AlexNet and VGG models with their numerous layers, SqueezeNet achieves AlexNet-level accuracy for image classification with significantly fewer parameters. Next, we picked DenseNets~\cite{densenet}, which are a type of CNN architecture known for their efficient use of parameters and strong feature propagation. We also selected GoogLeNet developed by researchers at Google and InceptionNet~\cite{inception}, which was built upon the success of GoogLeNet with more complex CNN architectures utilizing the Inception module as a core component. ShuffleNet~\cite{shufflenet} was also picked, which is a convolutional neural network architecture specifically designed for deployment on mobile and other resource-constrained devices. Next, we picked MobileNet~\cite{mobilenets}, which is a lightweight convolutional neural network architecture designed for mobile and embedded devices. Furthermore, we selected ResNeXt~\cite{resnext} that builds upon the success of ResNet (Residual Network) architecture, introducing a new concept called cardinality. Wide ResNet (WRN)~\cite{wideresnet} was another family we picked, which is a variant of the popular ResNet architecture specifically designed to address limitations associated with very deep networks. For automatic neural network architecture search applications on mobile devices, we picked MNASNet~\cite{mnasnet}--Mobile Neural Architecture Search Net. Finally, we picked EfficientNet~\cite{efficientnet}, which is a family of convolutional neural networks (CNNs) designed to achieve a balance between accuracy and efficiency.

\noindent\textbf{Baseline (without \sol):} For comparison, we adopted a baseline that executes multiple DNN models with a single script. From our observation, we realized that running separate processes for each model in the GPU is extremely inefficient because each process invokes a separate CUDA Context. Each CUDA Context takes about 500 MiB of additional GPU memory, consuming extra time and memory for each model. In order to present a fair comparison, we initiated multiple DNN models from a single script, which invokes a single CUDA Context for all of the models--ensuring optimal GPU memory consumption. Moreover, the script contains a trigger that checks for pending requests in the request queue. As soon as a model finishes its execution, the script checks for pending requests in the queue and initiates a new model if there is enough GPU memory. The Request Aggregator is responsible for aggregating requests from the users and putting them in the queue.

\noindent \textbf{Profiling Tools:} We used 2 different profiling tools to profile and collect data: Nvidia Nsight System and Pytorch Profiler.

\begin{figure}
    \centering
    \includegraphics[width=0.9\linewidth]{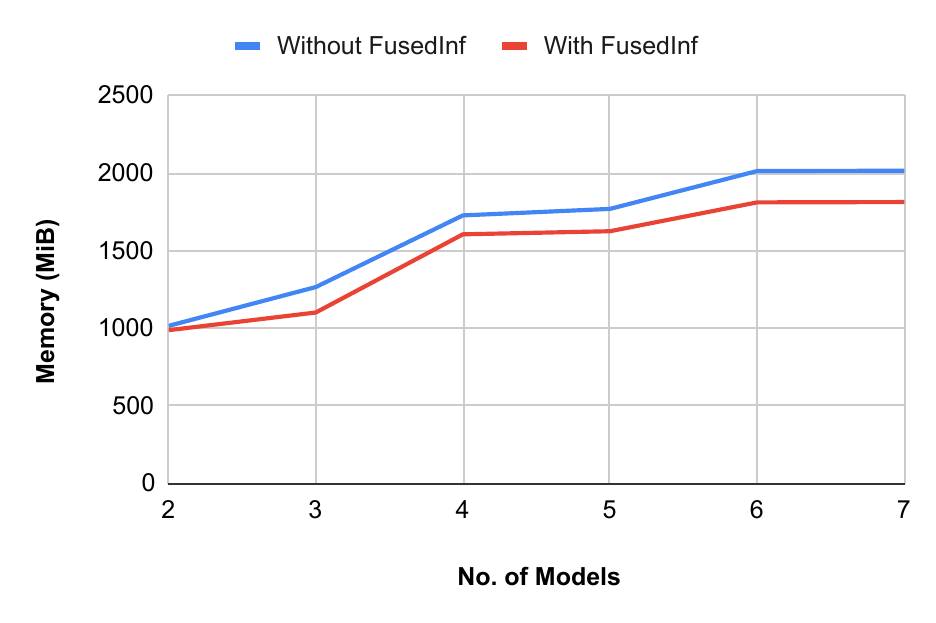}
    \vspace{-0.3cm}
        \caption{GPU memory consumption of different model combinations from 1 to 7 for 100 iterations.}
        \vspace{-0.3cm}
    \label{fig:7_model_result_mem}
\end{figure}

\begin{figure}
    \centering
    \includegraphics[width=0.9\linewidth]{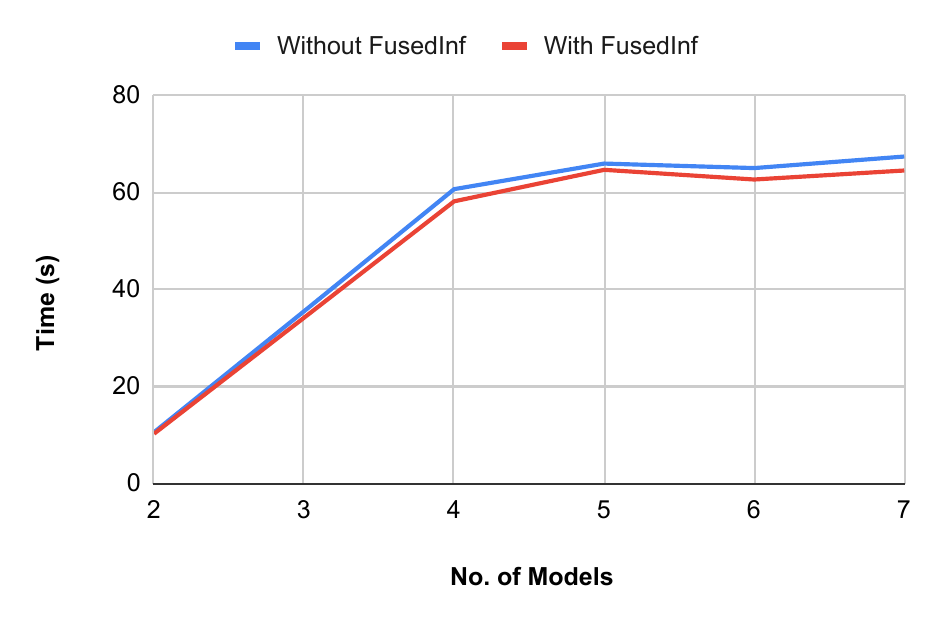}
    \vspace{-0.3cm}
    \caption{Total execution time of different model combinations from 1 to 7 for 100 iterations.}
    \label{fig:7_model_result}
    \vspace{-0.3cm}
\end{figure}

\subsection{Impact on Time and Memory with Different Number of Models}
\label{subsec:eval1}
\begin{figure}
    \centering
    \includegraphics[width=0.9\linewidth]{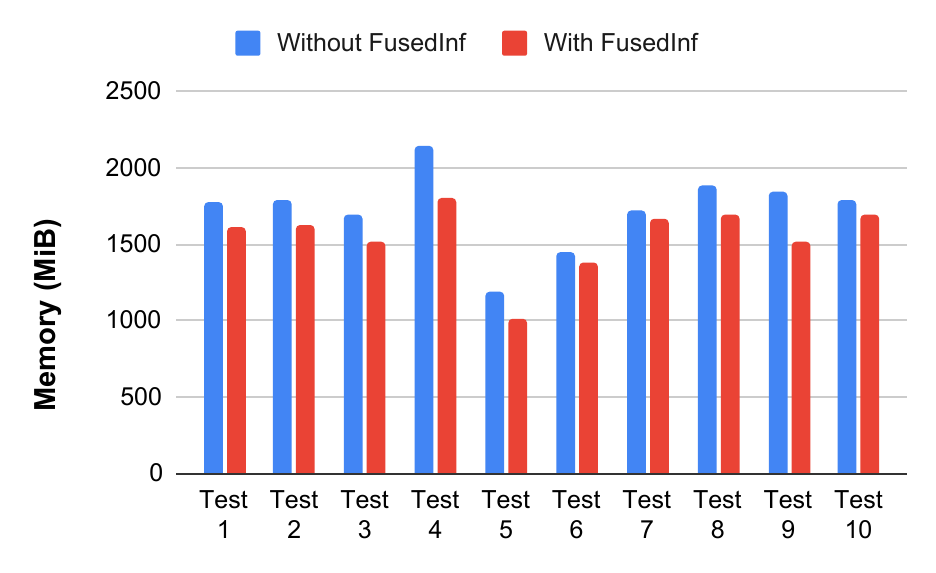}
    \vspace{-0.3cm}
        \caption{GPU memory consumption of 5 randomly picked models for 10 different test cases after 100 iterations each.}
        \vspace{-0.3cm}
    \label{fig:5_model_result_mem}
\end{figure}

\begin{figure}
    \centering
    \includegraphics[width=0.9\linewidth]{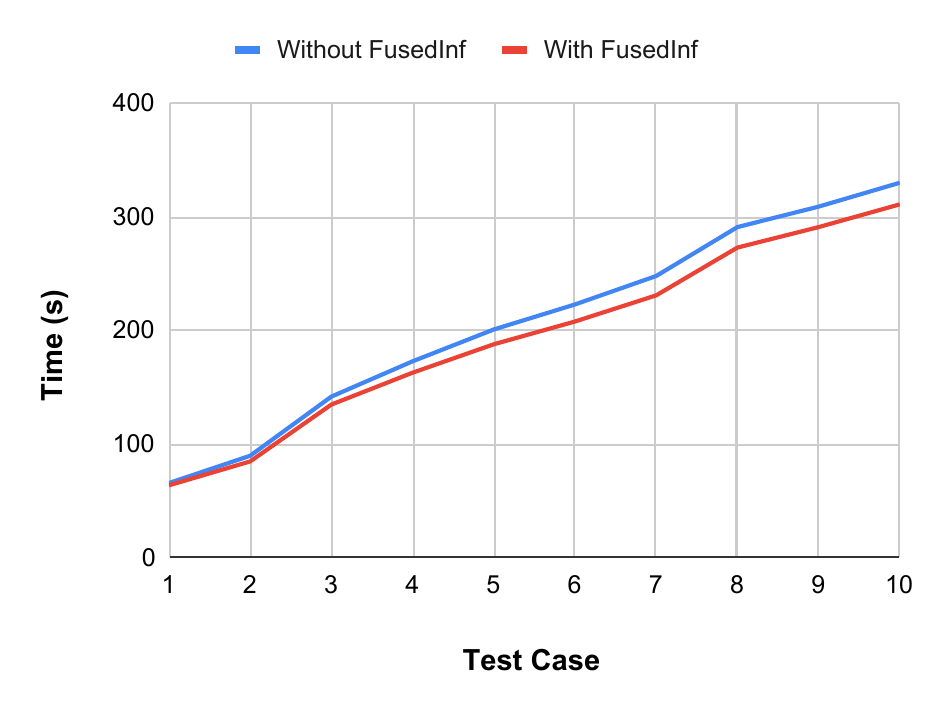}
    \vspace{-0.3cm}
    \caption{Total execution time of 5 randomly picked models for 10 different test cases after 100 iterations each.}
    \vspace{-0.3cm}
    \label{fig:5_model_result}
\end{figure}

In the first phase, we experimented with 7 DNN models. We gradually increased to 7 DNN models starting from 1 DNN model. As such, we ran 7 different experiments for this phase, each with 100 iterations per model. The size of the inputs for the models was (3, 224, 224)--images of 3 channels with a dimension of 224$\times$224. Initially, we picked the VGG16 model with batch normalization. When we ran 100 iterations of this single model, it consumed 954 MiB of GPU memory and the total time was 4.3 seconds. Next, we ran VGG16 with MobileNetV3 large. Without \sol, the two models peaked at 1016 MiB in GPU memory usage and the total execution time was 10.57 seconds. For the two models with \sol, the peak GPU memory usage was 988 MiB and the total execution time was 10.26 seconds. \sol was able to save 28 MiB (2.8\% less) of GPU memory and 0.31 seconds (3\% less) of execution time with two DNN models. Afterward, we ran VGG16, MobileNetv3Large, and DenseNet161 together. With these three models, the peak GPU memory usage was 1266 MiB without \sol and 1102 MiB with \sol. This time, \sol saved 164 MiB of GPU memory and 1.33 seconds of execution time with three DNN models. In the next run, we ran VGG16, MobileNetv3Large, DenseNet161, and EfficientNetV2Large together. This run peaked at 1728 MiB in GPU memory usage and 1606 MiB GPU memory usage without and with \sol, respectively. In terms of total execution time, the time was 60.71 seconds and 58.2 seconds, respectively. As such, \sol saved 122 MiB (7.6\% less) of GPU memory and 2.51 seconds (4.3\% less) of execution time with 4 DNN models. For the fifth test, we included ResNet18 with the rest of the 4 DNN models--running VGG16,
MobileNetv3Large, DenseNet161, EfficientNetV2Large, and ResNet18 at the same time. When the system was running there 5 DNN models, the peak GPU memory usage was 1770 MiB without \sol and 1626 MiB with \sol, which saved 144 MiB of GPU memory (a reduction of 9\%). In terms of execution time, it took 66 seconds without \sol and 64.7 seconds with \sol--a reduction of 1.29 seconds (2\% less) with 5 DNN models. The following run included AlexNet to the rest of the 5 DNN models, making a combination of 6 DNN models. For this run, the GPU memory consumption was 2014 MiB without \sol and 1812 MiB with \sol. As such, \sol saved 202 MiB of GPU memory (11.1 \% less). When running 6 DNN models at the same time, the execution time without \sol was 65.06 seconds, and with \sol, it was 62.7 seconds--reducing the time by 2.36 seconds (3.8\% less) for 6 DNN models. Finally, we ran 7 DNN models (VGG16, MobileNetv3Large, DenseNet161, EfficientNetV2Large, ResNet18, AlexNet, and SqueezeNet). Without \sol, the 7 DNN models used 2016 MiB of GPU memory, and the execution time was 67.44 seconds. With \sol, the GPU memory usage went down to 1814 MiB and the execution time was 64.56 seconds. As such, \sol was able to save 202 MiB of GPU memory (11.1\% less), and the execution time was 2.88 seconds quicker (4.5\% less). Figure~\ref{fig:5_model_result_mem} and \ref{fig:5_model_result} present the GPU memory consumptions and total execution time of this phase, respectively.

\subsection{Impact on Time and Memory with Consecutive Swaps}
\label{subsec:eval2}
In this next phase of experiments, we randomly picked 5 models out of the 34 DNN models belonging to 13 different model families. We ran 10 consecutive test cases randomly picking 5 DNN models each time and swapping them. This evaluation presents the effectiveness of the scheduler that swaps out DAGs after a fixed number of interactions (100 in this case).

We collected the same metrics as we did for the first phase. The first test case was inducted with AlexNet, VGG16 (with batch normalization), ResNet18, DenseNet161, MobileNetV2, and EfficientNetV2 Large. This test case demonstrated a reduction of 164 MiB in peak GPU memory usage (9.27\% less) and 2 seconds faster execution (3.13\% less). The second test case was inducted with AlexNet, VGG11, VGG19, ResNet50, MobileNetV3 Large, and Wide ResNet-50-2. This test case demonstrated a reduction of 160 MiB in peak GPU memory usage (8.96\% less) and 3 seconds faster execution (14.3\% less). In the third case, AlexNet, VGG13 (with batch normalization), ResNet34, Densenet-201, and ShuffleNet V2 x1.0 were picked randomly. This test case demonstrated a reduction of 172 MiB in peak GPU memory usage (10.17\% less) and 2 seconds faster execution (4\% less). The fourth test case was randomly picked VGG-11, VGG-19 (with batch normalization), ResNet-34, ResNet-152, and ShuffleNet V2 x1.0. This test case demonstrated a reduction of 344 MiB in peak GPU memory usage (16.03\% less) and 3 seconds faster execution (10.71\% less). In the fifth case, we randomly picked ResNet-34, SqueezeNet 1.1, ShuffleNet V2 x0.5, MobileNetV2, and Wide ResNet-101-2. This test case demonstrated a reduction of 174 MiB in peak GPU memory usage (14.6\% less) and 3 seconds faster execution (12\% less). The sixth test case included VGG19, ResNet-18, ResNet-50, MobileNetV2, and Wide ResNet-50-2. This test case demonstrated a reduction of 66 MiB in peak GPU memory usage (4.56\% less) and 2 seconds faster execution (10\% less). The seventh case was inducted AlexNet, VGG11 (with batch normalization), ResNet-101, SqueezeNet 1.0, and ResNeXt-101-32x8d. This test case demonstrated a reduction of 58 MiB in peak GPU memory usage (3.36\% less) and 2 seconds faster execution (8.7\% less). The eighth case included VGG16, ResNet-101, ResNet-152, ResNeXt-101-32x8d, and MNASNet 1.0 randomly. This test case demonstrated a reduction of 198 MiB in peak GPU memory usage (10.47\% less) and 1 second faster execution (2.38\% less). The following test case was conducted with AlexNet, VGG11, ResNet-34, Wide ResNet-50-2, and MNASNet 1.0. This test case demonstrated a reduction of 324 MiB in peak GPU memory usage (17.63\% less) but about the same execution time. In the final test, VGG13, VGG19, MobileNetV2, MobileNet V3 Large, and ResNeXt-50-32x4d were randomly chosen. This test case demonstrated a reduction of 92 MiB in peak GPU memory usage (5.14\% less) and 1 second faster execution (5\% less).

Figure~\ref{fig:5_model_result_mem} and \ref{fig:5_model_result} presents the peak GPU memory consumption and the total execution time for the consecutive test cases, receptively. On average, \sol saves 2 seconds per 30 seconds of operation executing 5 models. After 10 consecutive runs with 5 different models, the total execution time with \sol was 311 seconds, compared to 330 seconds without \sol. Following this trend, service providers will be able to save about 3 hours and 12 minutes per day, allowing them to run approximately 2000 more models per day.

\subsection{Impact on Time and Memory with Individual Model Swaps in a DAG}

\begin{table*}
\small
\caption{Memory and time of different swaps within a DAG.}
\label{tab:dag_swap}
\centering
\begin{tabular}{|c|l|l|l|l|l|l|}
\hline
\multirow{6}{*}{\makecell{Models\\in\\DAG}}                  & \textbf{Swap 1}               & \textbf{Swap 2}               & \textbf{Swap 3}               & \textbf{Swap 4}               & \textbf{Swap 5}               & \textbf{Swap 6}               \\ \cline{2-7} 
                                                         & VGG-19 with BN                & \textit{EfficientNet V2}    & \textit{EfficientNet V2}    & \textit{EfficientNet V2}    & \textit{EfficientNet V2}    & \textit{EfficientNet V2}    \\ \cline{2-7} 
                                                         & ResNet-50                     & ResNet-50                     & \textit{Inception v3}         & \textit{Inception v3}         & \textit{Inception v3}         & \textit{Inception v3}         \\ \cline{2-7} 
                                                         & MobileNet V3 L            & MobileNet V3 L                & MobileNet V3 L                & MobileNet V3 L                & \textit{VGG-16}               & \textit{VGG-16}               \\ \cline{2-7} 
                                                         & ResNeXt-50-32x4d              & ResNeXt-50-32x4d              & ResNeXt-50-32x4d              & \textit{SqueezeNet 1.1}       & \textit{SqueezeNet 1.1}       & \textit{SqueezeNet 1.1}       \\ \cline{2-7} 
                                                         & MNASNet 1.0                   & MNASNet 1.0                   & MNASNet 1.0                   & MNASNet 1.0                   & MNASNet 1.0                   & \textit{WideResNet-101}    \\ \hline
\multicolumn{1}{|l|}{\makecell{Memory\\without\\\sol\\(MiB)}}   & \multicolumn{1}{c|}{1384} & \multicolumn{1}{c|}{1268} & \multicolumn{1}{c|}{1264} & \multicolumn{1}{c|}{1180} & \multicolumn{1}{c|}{1680} & \multicolumn{1}{c|}{2168} \\ \hline
\multicolumn{1}{|l|}{\makecell{Memory\\with\\\sol\\(MiB)}}      & \multicolumn{1}{c|}{\makecell{1186\\(-16.69\%)}} & \multicolumn{1}{c|}{\makecell{1114\\(-13.82\%)}} & \multicolumn{1}{c|}{\makecell{1112\\(-13.67\%)}} & \multicolumn{1}{c|}{\makecell{1026\\(-15.01\%)}} & \multicolumn{1}{c|}{\makecell{1528\\(-9.95\%)}} & \multicolumn{1}{c|}{\makecell{2016\\(-7.54\%)}} \\ \hline
\multicolumn{1}{|l|}{\makecell{Time\\without\\\sol\\(s)}} & \multicolumn{1}{c|}{24.23}    & \multicolumn{1}{c|}{68.18}    & \multicolumn{1}{c|}{125.16}   & \multicolumn{1}{c|}{175.14}   & \multicolumn{1}{c|}{223.25}   & \multicolumn{1}{c|}{275.66}   \\ \hline
\multicolumn{1}{|l|}{\makecell{Time\\with\\\sol\\(s)}}    & \multicolumn{1}{c|}{\makecell{23.43\\(-3.41\%)}}    & \multicolumn{1}{c|}{\makecell{66.11\\(-3.13\%)}}    & \multicolumn{1}{c|}{\makecell{120.25\\(-4.08\%)}}   & \multicolumn{1}{c|}{\makecell{169.44\\(-3.36\%)}}   & \multicolumn{1}{c|}{\makecell{215.54\\(-3.58\%)}}   & \multicolumn{1}{c|}{\makecell{266.9\\(-3.28\%)}}    \\ \hline
\end{tabular}
\vspace{-0.3cm}
\end{table*}
\label{subsec:eval3}
In the final phase, we evaluated the sub-graph (model) swapping feature of \sol. We started with 5 different models and swapped each after completing the required iterations. This evaluation presents the effectiveness of the dynamic adaptive DAG compiler. \sol is capable of compiling a part of the DAG without the need to recompile the entire DAG.

The initial DAG contained VGG-19 with Batch Normalization, ResNet-50, MobileNet V3 Large, ResNeXt-50 (32x4d), and MNASNet 1.0. The first 25 iterations were 0.8 seconds faster (3.42\%) consuming 198 MiB less (16.69\%) GPU memory. After the 25th iteration, we swapped VGG-19 with EfficientNet where \sol was ahead by 2.07 seconds (3.13\%) consuming 154 MiB less memory (13.82\%). After 50 iterations, we swapped ResNet-50 with Inception v3, where \sol was ahead by 4.91 seconds (4.08\%) consuming 152 MiB less memory (13.67\%). After 75 iterations, we swapped ResNeXt-50-32x4d with SqueezeNet 1.1, where \sol was ahead by 5.7 seconds (3.36\%) consuming 154 MiB less memory (15.01\%). After 100 iterations, we swapped MobileNet V3 Large with VGG-16, where \sol was ahead by 7.71 seconds (3.58\%) consuming 152 MiB less memory (9.95\%). Finally, after 125 iterations, we swapped MNASNet 1.0 with Wide ResNet-101-2, and ran 25 iterations, where \sol was ahead by 8.76 seconds (3.28\%) consuming 152 MiB less memory (7.54\%).

Table~\ref{tab:dag_swap} presents the results after each swap. The GPU memory presents the peak consumption after each swap. The time presented is the total execution time after each swap. From the results, we can see that \sol was able to save an average of 160 MiB of memory (12.05\%) and a total of 8.76 seconds (3.28\%) of total execution time. 

\section{Discussion}
\sol is designed to facilitate DNN model swapping on resource-constrained edge boxes. From the evaluation results, we can see that \sol is capable of efficiently swapping DNN models in various use cases. The evaluation in Section~\ref{subsec:eval1} demonstrates that the greater the number of models, the more efficient \sol can be in terms of execution time and memory consumption. The evaluation in Section~\ref{subsec:eval1} shows that the total execution time with \sol was 19 seconds faster after 10 consecutive runs with 5 different models running for 100 iterations per cycle. This translates to 3 hours and 12 minutes saved per day, allowing a service provider to perform approximately 2000 more swaps per day, which is a significant number. Finally, Section~\ref{subsec:eval3} demonstrates the efficiency of swapping individual DNN models within a compiled DAG. \sol can not only efficiently compile a DAG of multiple models but also recompile a DAG efficiently by swapping a particular sub-graph (model) from the DAG.

\section{Related Work}

There are a few production-level inference systems offered by popular machine learning frameworks. TorchServe is one of the most popular inference engine services provided by PyTorch\cite{pytorch}. TorchServe is a flexible and high-performance tool designed specifically for serving PyTorch deep learning models in production environments. TorchServe provides a built-in web server that allows applications to make predictions using the deployed model through a set of easy-to-use REST APIs. TensorFlow serving is another tool by TensorFlow~\cite{tensorflow2015-whitepaper} for deploying inference engines on the edge. It is a software library designed specifically for deploying machine learning models trained with TensorFlow in production environments on the edge. Moreover, it provides both RESTful API and gRPC interfaces for clients to interact with the models, making it accessible to various edge development environments. ONNX, short for Open Neural Network Exchange,  functions as an open-source format for representing machine learning models. It acts as a common language, enabling a seamless exchange of models between different deep-learning frameworks. This interoperability empowers developers to train models in their preferred framework (like TensorFlow or PyTorch) and then deploy them on various edge platforms or runtimes that support ONNX~\cite{onnx}. Clipper is a low-latency online prediction serving system proposed by Crankshaw et. al.~\cite{clipper}. Clipper is a system designed to take machine learning models from various frameworks and optimize their performance for real-world use. It sits between applications and the models, simplifying deployment and using techniques like caching and batching to deliver predictions faster and more accurately.

Popular cloud service providers are also allowing users to deploy and run DNN models on their cloud or edge infrastructure by providing their own services. AWS SageMaker~\cite{aws_sagemaker} is a cloud-based platform offered by Amazon Web Services (AWS) specifically designed to streamline the ML workflow. It simplifies the process of building, training, deploying, and managing ML models on the edge. Azure Machine Learning (Azure ML~\cite{barnes2015azure}) is a cloud-based service offered by Microsoft. It streamlines the entire machine learning lifecycle, from data preparation and model training to deployment and monitoring on the edge. Vertex AI~\cite{vertex_ai} by Google Cloud simplifies the AI development process on the edge by providing a central platform for data management, model training, and deployment.

In 2021, Romero et. al. proposed INFaaS~\cite{infaas}--an automated model-less inference serving system. INFaaS simplifies deploying DNN models for real-time use. Instead of developers choosing specific DNN models for each task, INFaaS automatically selects the best option based on the desired performance (latency) and accuracy trade-off. applications send their requests to INFaaS through a user-friendly interface (Front-End). The core system (Controller) then analyzes these requests and picks the most fitting model variant for the job. This chosen variant, along with the actual query, is then directed to a Worker machine. Finally, the Worker leverages the appropriate hardware components (Hardware Executors) to execute the inference task and sends the results back to the application. However, in its current implementation, a Worker is only responsible for handling queries of one DNN model, which is not very efficient for the resource-contained edge devices.

AWS SageMaker recently introduced serverless inference services~\cite{serverless-sagemaker} for deploying DNN models that automatically scale resources based on incoming requests. This eliminates the need for manual server management and is ideal for workloads with unpredictable traffic patterns, as it only allocates resources when needed. This saves costs compared to constantly running servers and simplifies deployment for small to medium-sized businesses. Vertex AI Pipeline~\cite{serverless-vertex} by Google Cloud offers similar functionalities. Vertex AI Pipelines eliminates the need to manage machine learning projects and allows users to build automated workflows that handle everything from training the DNN models to monitoring their performance--all without needing to manage servers.

AMPS-inf~\cite{amps-inf} is a framework developed to exploit the serverless inference services to mitigate the management and overall cost while meeting the response time. AMPS-inf achieves that by using a technique of model partitioning. This involves breaking down the model into smaller pieces that can be run independently. It then formulates a Mixed-Integer Quadratic Programming problem to determine how many partitions to split the model into and how to assign those partitions to be run on serverless functions. The appropriate resources for each serverless function (e.g., memory, CPU) are acquired by solving this problem, AMPS-Inf aims to find the most cost-effective way to run the inference tasks while still meeting the required response time.

Ali et. al. proposed a framework called BATCH~\cite{batch} for latency performance and cost-effectiveness of machine learning inference. BATCH uses an optimizer to provide inference tail latency guarantees and cost optimization and to enable adaptive batching support. To meet the service level objectives BATCH adopts adaptive parameter tuning, which allows it to dynamically adjust the batching parameter based on the objectives defined by the user.

TETRIS is a serverless platform specifically designed for running deep learning inference tasks efficiently. It tackles the common problem of high memory usage in serverless environments by using a combination of techniques. TETRIS automatically shares resources like the tensors used by different inference tasks and reclaims unused memory and schedules serverless instances efficiently to minimize wasted resources. It also ensures the required performance standards of a serverless inference system.

\section{Conclusion}
\sol is designed to efficiently load and query multiple DNN models concurrently at the same time. The purpose of \sol is to facilitate the serverless inference services on the edge in order to serve more users per day by saving time when initializing and executing DNN models. The proposed framework achieves this by efficiently compiling a single DAG of multiple DNN models, which requires fewer memory calls, accelerates memory allocation, and achieves higher throughput. As such, \sol can allow a serverless inference service provider to serve more users at a time when there is a high traffic of queries for a wide variety of different DNN models. In the future, we will experiment with more model architectures and try to employ opportunistic approaches to find the best model combination for the most efficient utilization of GPU memory and function calls.

\bibliographystyle{IEEEtran}
\bibliography{ref}

\end{document}